\documentclass[sigconf]{acmart}

\usepackage{algorithm}
\usepackage{algorithmic}
\usepackage{graphicx}
\usepackage{subcaption}
\usepackage{xcolor}
\usepackage{caption}
\usepackage{tikz}
\usetikzlibrary{calc, positioning, fit, matrix}
\usetikzlibrary{shapes.geometric, arrows}

\tikzstyle{embedding} = [rectangle, 
minimum width=3cm, 
minimum height=1cm,
inner sep=5pt,
text centered, 
draw=black, 
fill=red!30]

\tikzstyle{mlp} = [rectangle, rounded corners,
minimum width=3cm,
minimum height=1cm,
inner sep=5pt,
text centered,
draw=black,
fill=darkgray!50]

\tikzstyle{attn} = [rectangle, rounded corners,
minimum width=3cm,
minimum height=1.25cm,
inner sep=5pt,
text width=3cm,
text centered,
draw=black,
fill=cyan!50]

\tikzstyle{gnn} = [rectangle, rounded corners,
minimum width=4cm, 
minimum height=1.5cm, 
text centered, 
text width=4cm, 
draw=black, 
fill=orange!30]

\tikzstyle{gru} = [rectangle, rounded corners,
minimum width=3cm, 
minimum height=1cm, 
text centered, 
text width=3cm, 
draw=black, 
fill=green!30]

\tikzstyle{concat} = [circle,
minimum size=0.5cm,
text centered, 
text width=0.5cm, 
draw=black, 
fill=blue!30]
\tikzstyle{arrow} = [thick,->,>=stealth]

\AtBeginDocument{%
  }

\setcopyright{acmlicensed}
\copyrightyear{2024}
\acmYear{2024}
\acmDOI{XXXXXXX.XXXXXXX}

\acmConference[CODS-COMAD Dec '24]{8th International Conference on Data Science and Management of Data (12th ACM IKDD CODS and 30th COMAD)}{December 18--21, 2024}{Jodhpur, India}
\acmBooktitle{8th International Conference on Data Science and Management of Data (12th ACM IKDD CODS and 30th COMAD) (CODS-COMAD Dec '24), December 18--21, 2024, Jodhpur, India}
\acmISBN{978-1-4503-XXXX-X/18/06}

\begin{document}

\title{Spatio-Temporal Forecasting of PM$_{2.5}$ via Spatial-Diffusion guided Encoder-Decoder Architecture}


\author{Malay Pandey $^1$, Vaishali Jain $^2$, Nimit Godhani $^2$, Sachchida Nand Tripathi $^{2, 3}$, Piyush Rai $^1$}
\affiliation{%
  \institution{$^1$ CSE Department, IIT Kanpur, $^2$ Civil Engineering Department, IIT Kanpur, $^3$ SEE Department, IIT Kanpur}
  \state{Uttar Pradesh}
  \country{India}
}





\renewcommand{\shortauthors}{}

\begin{abstract}
    In many problem settings that require spatio-temporal forecasting, the values in the time-series not only exhibit spatio-temporal correlations but are also influenced by spatial diffusion across locations. One such example is forecasting the concentration of fine particulate matter (PM$_{2.5}$) in the atmosphere which is influenced by many complex factors, the most important ones being diffusion due to meteorological factors as well as transport across vast distances over a period of time. We present a novel Spatio-Temporal Graph Neural Network architecture, that specifically captures these dependencies to forecast the PM$_{2.5}$ concentration. Our model is based on an encoder-decoder architecture where the encoder and decoder parts leverage gated recurrent units (GRU) augmented with a graph neural network (TransformerConv) to account for spatial diffusion. Our model can also be seen as a generalization of various existing models for time-series or spatio-temporal forecasting. We demonstrate the model's effectiveness on two real-world PM$_{2.5}$ datasets: (1) data collected by us using a recently deployed network of low-cost PM$_{2.5}$ sensors from 511 locations spanning the entirety of the Indian state of Bihar over a period of one year, and (2) another publicly available dataset that covers severely polluted regions from China for a period of 4 years. Our experimental results show our model's impressive ability to account for both spatial as well as temporal dependencies precisely. The code is publicly available at \url{https://github.com/malayp717/pm2.5}.
\end{abstract}



\maketitle

\section{Introduction}

Spatio-temporal forecasting~\cite{jin2023spatio} aims to integrate spatial and temporal information in data to predict future events or trends, and has widespread usage in diverse areas, such as climate science, urban planning, and epidemiology. The problem of spatio-temporal forecasting distinguishes itself from standard time-series forecasting as it needs to account for the evolution over time and across different locations. Unlike standard time-series forecasting, spatio-temporal forecasting requires capturing the complex interactions arising due to both space and time.

A motivating problem we consider in this work is from the domain of Air Quality Monitoring (AQM)~\cite{IQAir2021}; in particular, forecasting the concentration levels of fine particulate matter PM$_{2.5}$ (particles with aerodynamic diameter $\leq \, 2.5\, \mu m$ ) for a given set of stations at various geographical locations. Elevated levels of PM$_{2.5}$ pollution in the atmosphere have far-reaching health implications, including the development of severe medical conditions, such as cardiovascular obstructive pulmonary disease, lung cancer, stroke, and asthma \cite{JTD6353}. Accurate forecasts can help in taking mitigating measures and help inform policy-makers for early warning systems as well as long-term planning.

PM$_{2.5}$ concentration is characterized by complex processes, starting from emission generated by pollution sources to transport and diffusion influenced by meteorological and geographical information, which often have wide range and long-lasting effects. As reported by \cite{HU2014598} PM$_{2.5}$ can be transported hundreds of kilometers in 72 hours. Thus, not only does spatial diffusion play a significant role in PM$_{2.5}$ concentration, but long-term dependencies also play a role, making it a complex spatio-temporal forecasting problem.

For the forecasting of standard time-series data, a number of approaches have been considered in prior work, ranging from classical time-series forecasting methods, such as  autoregressive (AR) and exponential smoothing\cite{liu2021forecast}, to more recent deep learning approaches, such as recurrent neural networks and temporal convolutional neural networks~\cite{lim2021time}. However, while being effective for the forecasting of a single time-series, these methods are not able to leverage the complex dependencies across different locations, each of which is defined by its own time-series. This necessitates time-series forecasting methods that can account for spatial dependence across locations and has led to significant research interest in spatio-temporal forecasting~\cite{jin2023spatio}.

Spatio-temporal forecasts are often also dependent on not just the previous values in the time-series and the values at other neighboring locations, but also on exogeneous variables. For example, in the PM$_{2.5}$ forecasting problem, the concentrations of PM$_{2.5}$ at any station not only exhibit spatial and temporal dependencies but are also dependent on meteorological and other weather-related variables at that location. If thinking of each location as a node of a graph, these variables serve as node features. Effectively leveraging these diverse types of node features in the forecasting model thus becomes essential. The forecasting models need to take these variables into account to effectively capture spatial correlations, and phenomena such as transport and diffusion.

We present a spatio-temporal sequence-to-sequence model for the forecasting of PM$_{2.5}$ concentration levels at different locations. Our model is based on an encoder-decoder architecture where the sequence history is input to the encoder and the forecast values are the outputs of the decoder. While encoder-decoder based architectures have been considered in other works on time-series forecasting of PM$_{2.5}$, our work has several distinguishing aspects: (1) It augments the encoder-decoder framework with a graph neural network (GNN) to capture the spatial dependencies across locations and an attention mechanism to capture short as well as long-term temporal dependencies, and (2) The graph construction leverages various meteorological and weather-related variables to model the spatial-diffusion~\cite{Wang_2020}, a process which is known to contribute the the concentration levels of pollutants, such as PM$_{2.5}$. 

Graph Neural Networks (GNNs) are especially appealing in spatio-temporal forecasting problems because of their remarkable capability in effectively capturing and analyzing topological information. GNNs can represent the region of interest as a collection of nodes and edges. Nodes often represent meteorological data at specific locations, while edges indicate the interactions between distinct nodes. Due to the spatial and temporal characteristics of the problem, GNNs are frequently coupled with modifications to Recurrent Neural Networks (RNNs) to capture the temporal trends.  One such example is \cite{TENG2023107971} who suggested representing cities as nodes and constructing a weighted undirected graph that reflects the distance between these nodes. The GNN's message-passing mechanism transmits the outputs to Long-Short Term Memory (LSTM) for temporal modeling, culminating in the final predictions. However, this model does not account for spatial-diffusion.  

Our model is similar in spirit to~\cite{Wang_2020} who emphasized the need to include meteorological elements such as wind speed and wind direction as edge features, as they greatly affect the movement of fine-particulate materials over long distances. While the work of ~\cite{Wang_2020} does consider spatial diffusion, it has some key differences from our work: (1) The problem setting considered in \cite{Wang_2020} assumes that the attributes (meteorological and weather-related variables) at each location are also known (via a separate forecasting model) for future time-steps at which forecasts are to be made. This is a strong assumption which our model does not make; and (2) Our model makes use of attention mechanism on the decoder side (using Luong Attention) to capture short and long-range temporal dependencies whereas the model of \cite{Wang_2020} uses a standard GRU for the temporal modeling. These aspects make our model considerably more flexible, and more generally applicable as compared to methods such as~\cite{Wang_2020}, and the model architecture to also differ significantly from ~\cite{Wang_2020}.

We apply our spatial-diffusion guided
encoder-decoder model on two datasets: (1) A real-world PM$_{2.5}$ concentration dataset that we have collected from a recently deployed dense network of low-cost PM$_{2.5}$ sensors from 511 locations spanning the entirety of the Indian state of Bihar, over a period of one year (01/05/2023 - 30/04/2024), and (2) publicly available dataset that covers severely polluted areas in China, spanning over a period of four years (01/01/2015 - 31/12/2018)\footnote{\href{https://github.com/shuowang-ai/PM2.5-GNN}{KnowAir Dataset for severely polluted cities in China}}. In addition to our work proposing a spatio-temporal forecasting model that is novel in its own right, we apply this model, to the best of our knowledge, for the first such study of such a scale and nature in India.

Section~\ref{sec:method-dataset} introduces the dataset used in our study, followed by a description of the problem setting in Section~\ref{method:problem}, and our proposed model in Section~\ref{sec:method-architecture}. Section~\ref{sec:relwork} discusses related work and Section~\ref{sec:exp} presents our experimental results where we compare our model with several state-of-the-art methods in various settings.

\section{Methodology}
\label{sec:method}
We start by describing the dataset and the area of interest where our proposed model is evaluated. Subsequently, we formulate the PM$_{2.5}$ forecasting problem in mathematical terms and discuss construction of the graph over locations from our area of interest. Finally, we present the proposed architecture to model the spatio-temporal evolution of PM$_{2.5}$ concentration levels over the area of interest.

\subsection{Dataset}
\label{sec:method-dataset}
This study focuses primarily on the region of Bihar, India. An extensive network of Air Quality Monitoring (AQM) stations distributed across the entire state provided verifiable ground truth PM$_{2.5}$ concentrations, as well as readings for meteorological features, such as relative humidity (RH) and diurnal temperature (T), on an hourly basis. Please refer to Section \ref{sec:exp-dataset} to find the distribution of these AQM stations across the entire state.

Following \cite{Wang_2020}, we utilize data from two sources: the AQM stations (RH, T, PM$_{2.5}$) and the ERA5 climate reanalysis data from the European Centre for Medium-Range Weather Forecasts (ECMWF) \footnote{\href{https://cds.climate.copernicus.eu/datasets/reanalysis-era5-single-levels?tab=download}{ERA5 hourly data on single levels from 1940 to present}}, to obtain an additional set of features, namely: Planetary Boundary Layer (PBL) height, u10-component of wind, v-10 component of wind, K-index, Surface Pressure and Total Precipiation. We refer to the meteorological features and the features from ERA5 data collectively as "node attributes" which are available for each station. In addition to these node attributes, we also leverage space and time information; in particular, the location information (latitude-longitude) and time-stamp associated with the recorded PM$_{2.5}$ observations. The time-stamps (hour-of-the-day, day-of-the-week, and month) are important because PM$_{2.5}$ concentrations vary throughout the day, and the concentrations drop on the weekend. To address the issue of missing data (e.g., due to a malfunctioning sensor), we employed the Multiple Imputation by Chained Equations (MICE) algorithm \cite{mice} for missing data imputation.

To further examine the model's capabilities and robustness, we also leverage the KnowAir dataset, utilized by \cite{Wang_2020}, which covers the severely polluted regions of China. The dataset uses the same set of node attributes as the Bihar Dataset.

\subsection{Problem Definition}
\label{method:problem}
The PM$_{2.5}$ concentration forecasting can be formulated as a spatio-temporal problem. Let $y^{t} \in \mathbb{R}^{L}$ denote the PM$_{2.5}$ concentrations at time $t$, where $L$ is the total number of locations. We define a directed graph $G = (V, E)$, where $V$ is the set of nodes representing different locations, and $E$ is the set of edges denoting potential interactions among these locations. Let $X^{t} \in \mathbb{R}^{L \times  d}$ denote the node attributes for all locations at time-step $t$, where $d$ is the number of attributes/features available for each location at each time-step. Let us define $H$ and $F$ as the length of the history and forecasting periods, respectively. During the history period, we have access to the complete set of node attributes (denoted $X$), the location and time information (denoted by $\overline{X}$), and the ground truth PM$_{2.5}$ concentrations (denoted by $y$). During the forecasting period, however, we only have access to the location and time-stamp information ($\overline{X}$) for the PM$_{2.5}$ concentrations in our forecasting window.

Formally, we define our problem statement as follows: for any starting point $\tau$, we feed the observed node attributes ($X^{\tau}$), the ground truth PM$_{2.5}$ labels ($y^{\tau}$), and the graph structure $G$ for the history period of length $H$, along with the location and timestamp attributes ($\overline{X}^{\tau})$. The output of our model $f$ is the predicted PM$_{2.5}$ values for the specified forecasting period of length $F$. This can be mathematically represented as:
\begin{equation}
    \bigl[\, \hat{y}^{k+1} ..., \hat{y}^{T} \,\bigr] = f\bigl(\, X^{\tau} ..., X^{k};\, y^{\tau} ..., y^{k};\, \overline{X}^{\tau} ..., \overline{X}^{T};\, G \,\bigr)
\end{equation}
where $k = \tau+H$ and $T = \tau+H+F$.

It is important to highlight that, unlike prior work~\cite{Wang_2020}, we do not assume that the node attributes (Table 1) are available for the forecasting period. In~\cite{Wang_2020}, it was assumed that these attributes are provided using other forecasting methods. This can be an unrealistic assumption in practice.

For training, we minimize the error between predicted PM$_{2.5}$ values [$\hat{y}^{k+1}$ ..., $\hat{y}^{T}$], and ground truth [$y^{k+1}$ ..., $y^{T}$] using Mean Squared Error (MSE) Loss.

\begin{equation}
    \text{MSE Loss} = \frac{1}{L} \sum_{l=1}^{L} \, \biggl( \, \frac{1}{F} \sum_{t=k+1}^{T} \, \Bigl(\, \hat{y}_{l}^{t} - y_{l}^{t} \Bigr)^{2} \biggr)
\end{equation}

\begin{figure}[!htbp]
\begin{tikzpicture}[scale=0.60, node distance=2cm, every node/.style={transform shape}]
\node (embenc) [embedding] {[\,$X^{t} ..., X^{k};\, y^{t} ..., y^{k};\, G$\,]};
\node (gnn) [gnn, above of=embenc, yshift=0.5cm] {Graph Neural Network\\(TransformerConv)};
\node (concat) [concat, above of=gnn] {\Huge{+}};
\node (gruenc) [gru, above of=concat] {GRU};
\node (mlpenc) [mlp, above of=gruenc] {MLP};
\node (resultenc) [embedding, above of=mlpenc] {$[\,\hat{y}^{t} ..., \hat{y}^{k} \,]$};

\node [draw, dashed, fit=(gnn)(concat)(gruenc)(mlpenc)(resultenc), inner sep=0.4cm, label=above:\textbf{Encoder}] {};

\node (embdec) [embedding, right of=embenc, xshift=4cm] {[\,$\overline{X}^{k+1} ..., \overline{X}^{T};\, \hat{y}^{k}$\,]};
\node (grudec) [gru, above of=embdec, yshift=0.5cm] {GRU};
\node (attndec) [attn, above of=grudec] {Luong Attention};
\node (mlpdec) [mlp, above of=attndec] {MLP};
\node (resultdec) [embedding, above of=mlpdec] {[\,$\hat{y}^{k+1} ..., \hat{y}^{T}$\,]};

\node [draw, dashed, fit=(grudec)(attndec)(mlpdec)(resultdec), inner sep=0.4cm, label=above:\textbf{Decoder}] {};

\coordinate (P1) at ($(embenc.west) - (1.25,0)$);
\coordinate (P2) at ($(concat.west) - (1.25,0)$);
\coordinate (P3) at ($(resultenc.east) + (1.25,0)$);
\coordinate (P4) at ($(embdec.west) - (1.25,0)$);
\coordinate (P5) at ($(gruenc.east) + (1.25,0)$);
\coordinate (P6) at ($(attndec.west) - (1.25,0)$);

\draw [arrow] (embenc) -- (gnn);
\draw [arrow] (gnn) -- (concat);
\draw [arrow] (embenc.west) -- (P1) |- (P2) -- (concat.west);
\draw [arrow] (concat) -- (gruenc);
\draw [arrow] (gruenc) -- (mlpenc);
\draw [arrow] (mlpenc) -- (resultenc);
\draw [arrow] (resultenc.east) -- (P3) |- (P4) -- (embdec.west);
\draw [arrow] (embdec) -- (grudec);
\draw [arrow] (grudec) -- (attndec);
\draw [arrow] (gruenc.east) -- (P5) |- (P6) -- (attndec.west);
\draw [arrow] (attndec) -- (mlpdec);
\draw [arrow] (mlpdec) -- (resultdec);

\end{tikzpicture}
\caption{Architecture of the proposed AGNN\_GRU model where the encoder (left block) consists of a GNN (TransformerConv) and decoder (right block) consists of Luong Attention}
\label{fig:model}
\vspace{-2em}
\end{figure}
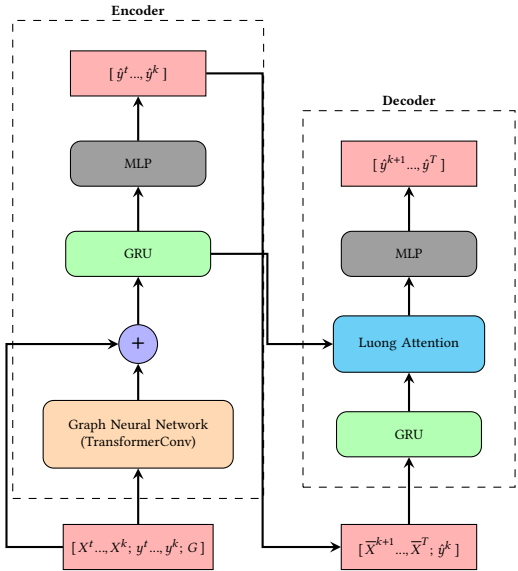

\subsection{Our Model: AGNN\_GRU}
\label{sec:method-architecture}

As illustrated in Figure \ref{fig:model}, the proposed architecture represents an enhanced version of the Sequence-to-Sequence model, consisting of two main components: an Encoder (left block of Figure \ref{fig:model}) and a Decoder (right block of Figure \ref{fig:model}). We refer to our model as AGNN\_GRU where `A' stands for Attention, specifically the Luong attention mechanism\cite{luong2015effectiveapproachesattentionbasedneural} employed in the decoder; `GNN' denotes the Graph Neural Network incorporated in the encoder to capture spatial dependencies, including spatial-diffusion; and `GRU' represents the Gated Recurrent Units used in both the encoder and decoder to model temporal sequences. This combination of GNN and GRU in the encoder, coupled with an attention-enhanced GRU in the decoder, forms a powerful framework for capturing both spatial and temporal dynamics in sequence-to-sequence tasks.

The primary objective of the encoder is to identify trends based on historical node attributes, location and time-stamp, and PM$_{2.5}$ concentrations. This is accomplished by initially capturing spatial dependencies through a Graph Neural Network (GNN) and subsequently integrating these to capture temporal dependencies using a Gated Recurrent Unit (GRU).
\vspace{-1mm}
\subsubsection{Graph Construction}

For directed graph construction, we follow a methodology similar to the one employed in \cite{Wang_2020}. In this approach, the nodes represent the meteorological characteristics of different locations in our dataset. The edges define how these nodes will interact with one another. The key idea behind edge attributes is influenced by wind speed and direction since it is a major contributor to the PM$_{2.5}$ horizontal transport. The edge attributes are defined by the following values: distance between source and sink, angle between source and sink, source wind speed, source wind direction, and Advection Coefficient.


Following the Transformer Convolution \cite{shi2021maskedlabelpredictionunified} message-passing paradigm, the GNN module of our encoder (left block of Figure \ref{fig:model}) learns enhanced spatial representations by iteratively aggregating neighboring information on the graph, as formulated in Equations \ref{transconv_start}-\ref{transconv_end}. Since the graph is directed, and the edges encapsulate the wind speed and direction information, the edge attributes help enhance the spatial representations by establishing a direct relationship between the source and sink. The Transformer Convolution is mathematically defined as

\begin{align}
\vspace{-2em}
\label{transconv_start}
    \eta_{i}^{t} &= W_{1} \cdot P_{i}^{t} + \sum_{j \in \mathcal{N}_{i}} \alpha_{ij}^{t} \Bigl( W_{2} \cdot P_{j}^{t} + W_{6} \cdot E_{i \rightarrow j}^{t} \Bigr)\\
    \alpha_{ij}^{t} &= \text{softmax} \biggl(\frac{\bigl(W_{3} \cdot P_{i}^{t}\bigr)^{\text{T}} \bigl( W_{4} \cdot P_{j}^{t} + W_{5} \cdot E_{i \rightarrow j}^{t} \bigr)}{\sqrt{d}} \biggr)
\label{transconv_end}
\vspace{-1em}
\end{align}
where $W_{1}, W_{2} \cdots W_{6}$ represent trainable parameters, that are shared across all time-stamps, $P_{i}^{t}$ represents meteorological, location, and air quality information at location $i$ and time-step $t$, as expressed in Equation \ref{enc_start}, $E_{i \rightarrow j}^{t}$ represent the edge attributes of the directed edge between source ($i$) and sink ($j$), at time step $t$, and $d$ is the normalization constant added in the denominator, to help combat the exploding or vanishing gradient problem.

To learn the temporal trends, we first concatenate the enhanced spatial representations with the input node attributes, at each time-step, and send them to a Gated Recurrent Unit (GRU) \cite{chung2014empiricalevaluationgatedrecurrent}, as formulated in Equations \ref{gru_emb} and \ref{gru_enc}. This allows our encoder unit to model temporal transport, while also considering spatial diffusion with the help of the GNN. These operations are summarized as
\begin{align}
\label{enc_start}
    P^{t}_{i} &= [X^{t}_{i}, \overline{X}^{t}_{i}, y^{t}_{i}] &\forall t \in [\tau, k]\\
    \eta^{t} &= \text{TransformerConv}(G(P^{t}, E^{t}))\\
\label{gru_emb}
    \zeta^{t}_{i} &= [P^{t}_{i}, \eta^{t}_{i}] &\forall i \in [1, L]\\
\label{gru_enc}
    H^{t}_{i} &= \text{GRU}(\zeta^{t}_{i})\\
    \hat{y}^{t}_{i} &= \text{MLP}(H^{t}_{i})
\label{enc_end}
\vspace{-1em}
\end{align}

The decoder utilizes the patterns identified by the encoder to generate accurate PM$_{2.5}$ predictions. Unlike~\cite{Wang_2020}, due to the absence of meteorological data during the entirety of the forecasting period, it is not reasonable to include GNN on the decoder side as we are unable to consider the horizontal movement of particulate matter. To address this, we utilize the location and time-stamp features and the latest output from encoder $\hat{y}^{k}$, which are then transmitted to a Gated Recurrent Unit (GRU) as described in Equations \ref{dec_start} and \ref{gru_dec} (also see Figure ~\ref{fig:model}). Next, in order to incorporate historical trends, we compute improved outputs from GRU by utilizing Luong Attention \cite{luong2015effectiveapproachesattentionbasedneural}, as shown in Equation \ref{attn_dec}. These enhanced outputs are then sent to a Multi-Layer Perceptron (MLP) to obtain the final normalized PM$_{2.5}$ predictions. These operations are summarized by the following equations
\begin{align}
\label{dec_start}
    Q^{t}_{i} &= [\overline{X}^{t}_{i}, \hat{y}^{t-1}_{i}] &\forall t \in [k+1, T]\\
\label{gru_dec}
    \eta^{t}_{i} &= \text{GRU}(Q^{t}_{i}) &\forall i \in [1, L]\\
\label{attn_dec}
    \pi^{t} &= \text{LuongAttention}(H, \eta^{t})\\
    \hat{y}^{t}_{i} &= \text{MLP}(\pi^{t}_{i})
\label{dec_end}
\end{align}

Note that, to model temporal dependencies, we further extend the Gated Recurrent Unit (GRU), to not only consider long temporal dependencies for each station, but also factor in the spatial dependencies at each time step. Specifically, we incorporate output from GNN at the encoder step and get enhanced outputs using Luong Attention (Equation~\ref{attn_dec}) at the decoder step. 

The various steps used by the GRU in the encoder and decoder size can be summarized by Equations \ref{gru_start}-\ref{gru_end}.

\begin{align}
\label{gru_start}
    x_{i}^{t} &= \begin{cases}
        [P_{i}^{t}, \eta_{i}^{t}] & \forall t \in [\tau, k]\\
        Q_{i}^{t} & \forall t \in [k+1, T]
    \end{cases}\\
    z_{i}^{t} &= \sigma \Bigl(W_{z} \cdot [h_{i}^{t-1}, x_{i}^{t}] \Bigr)\\
    r_{i}^{t} &= \sigma \Bigl(W_{r} \cdot [h_{i}^{t-1}, x_{i}^{t}] \Bigr)\\
    \tilde{h}_{i}^{t} &= \text{tanh} \Bigl(W_{\tilde{h}} \cdot [r_{i}^{t} \odot h_{i}^{t-1}, x_{i}^{t}] \Bigr)\\
    h_{i}^{t} &= (1 - z_{i}^{t}) \odot h_{i}^{t-1} + z_{i}^{t} \odot \tilde{h}_{i}^{t}
\label{gru_end}
\end{align}
where $W_z, W_r$ and $W_{\tilde{h}}$ represent trainable parameters, that are shared across all the locations, and $\odot$ represents Hadamard Product. The entire procedure for training can be summarized using the following pseudo-code.

\begin{algorithm}[H]
\caption{AGNN\_GRU model}
\label{alg:agnn_gru}
\begin{algorithmic}[1]
\REQUIRE
\STATE Observed Meteorological Attributes $[X^{\tau} \hdots\, X^{k}]$
\STATE Observed PM$_{2.5}$ values $[y^{\tau} \hdots\, y^{k}]$
\STATE Known Location and Timestamp Attributes $[\overline{X}^{\tau} \hdots\, \overline{X}^{T}]$
\STATE G = (V, E)
\ENSURE Predicted PM$_{2.5}$ values $[\hat{y}^{k+1} \hdots\, \hat{y}^{T}]$
\STATE Initialize H $\gets$ []
\FOR{$t = \tau \hdots\, k$}
    \STATE $h^{t}, \hat{y}^{t}$ = Encoder($X^{t}, y^{t}, \overline{X}^{t}, G)$ \hfill(Equations \ref{enc_start}-\ref{enc_end})
    \STATE Append $h^{t}$ to H
\ENDFOR
\STATE Initialize preds $\gets$ []
\FOR{$t = (k+1) \hdots\, T$}
    \STATE $\hat{y}^{t}$ = Decoder($\hat{y}^{t-1},\, \overline{X}^{t},\, H)$ \hfill(Equations \ref{dec_start}-\ref{dec_end})
    \STATE Append $\hat{y}^{t}$ to preds
\ENDFOR
\RETURN preds
\end{algorithmic}
\end{algorithm}

This architecture captures complex spatio-temporal relationships while remaining lighter than transformers due to the absence of self-attention and cross-attention mechanisms, reducing parameter size. While transformers excel with large datasets, their complexity may cause overfitting on shorter time-series data like ours.

\section{Related Work}
\label{sec:relwork}
The problem of spatio-temporal forecasting has been studied extensively in several prior works; see Z. Liu et al. \cite{jin2023spatio} for an excellent survey on this topic, \cite{zhou2024deep} for specifically for deep learning architectures for prediction of PM$_{2.5}$, and \cite{han2023machine} for machine learning methods for air-quality analytics in general. 

In this section, we mainly focus on forecasting of air-quality (e.g., fine particulate matter concentrations) using commonly used time-series methods as well as more sophisticated spatio-temporal approaches that take into account both spatial and temporal information in the air-quality data from multiple locations.

\subsection{Classical Time-Series Methods}

Time-series forecasting methods have been traditionally employed to predict air quality. These methods include classical statistical models like ARIMA \cite{ZHAO2022e12239}, which are designed to capture temporal dependencies in data. While these models are straightforward and interpretable, they often fall short when it comes to capturing the complex spatio-temporal dynamics present in air quality data.

Deep learning based sequential data models, such as RNN, LSTM, GRU, and their bi-directional counterparts have also been used for time-series forecasting~\cite{petry2021design,kim2023comparison}; however, these approach do not consider the spatial information.
\subsection{Spatio-Temporal Approaches}

MasterGNN~\cite{han2021joint} performs joint prediction of air-quality and weather variables using a multi-adversarial spatio-temporal network, while we focus on the forecasting of PM$_{2.5}$ using weather and other relevant variables from the historical data, and do not aim to forecast these weather variables which was the focus of ~\cite{han2021joint}. 

Dynamic directed spatio-temporal graph convolution networks (DD-STGCN) \cite{zhou2021forecasting} explicitly considers the effect of dynamic wind-field for predicting PM$_{2.5}$ at the next step $T$ given the graphs for all previous time-steps $1,2,\ldots,T-1$. This is a different setting than our as it does not consider meteorological variables and long-horizon forecasting. 

PM$_{2.5}$-GNN~\cite{Wang_2020} is the most closely related to our approach in how spatial-diffusion is taken into account in forecasting. However, a key difference between our work and \cite{Wang_2020} is that they assume access to the forecasts of meteorological variables whereas our method does not.

\subsection{Other Related Works}

Another recent method Airformer~\cite{liang2023airformer} leverages transformers for forecasting air-quality. While leveraging the power of transformer architectures and also latent variables to capture the uncertainty in air-quality data, this approach does not explicitly capture the spatial diffusion phenomenon as captured by our model.

Among other related works, denoising diffusion model based approaches~\cite{wen2023diffstg,yang2024survey} have also been proposed for spatio-temporal forecasting leveraging their impressive generative modeling capability. \cite{chen2023group} developed group-aware GNN for nationwide city air quality forecasting. \cite{mirzavand2023explainable} focus on standard GRU based encoder-decoder architectures but with a focus on explainability, in particular to identify key meteorological and temporal features that contribute to air-pollutants. It would be of interest to integrate such approaches into our proposed method which also leverages various meteorological features.

In another recent work, \cite{chen2024adaptive} proposed an adaptive adjacency matrix-based graph convolutional recurrent network for air quality prediction. \cite{hu2023graph} used graph neural process for forecasting along with uncertainty estimates of the forecasts.

Spatio-temporal data can also exhibit distribution shifts over long time horizons. Recent work~\cite{liu2024pm2} has looked at this challenging problem, and it would be of interest to extend our approach using ideas in these works to handle distribution shifts.

Recent works have also started exploring the use of transformers and, more generally, foundation models for time-series forecasting~\cite{liang2024foundation,miller2024survey,yeh2023toward}. While promising, the current models do not consider aspects such as spatial relationships among the different locations, or other domain-specific knowledge, such as spatial-diffusion. It would be of interest to extend these models to take such information into account.

\section{Experiments}
\label{sec:exp}
In this section, we carry out comprehensive experiments to showcase the effectiveness of our model. Subsequently, we carefully analyze our model's performance, not only on a macroscopic scale but also region-wise, to demonstrate its robustness.

\subsection{Dataset}
\label{sec:exp-dataset}

\begin{figure}[!htbp]
    \centering
    \includegraphics[width=0.4\textwidth]{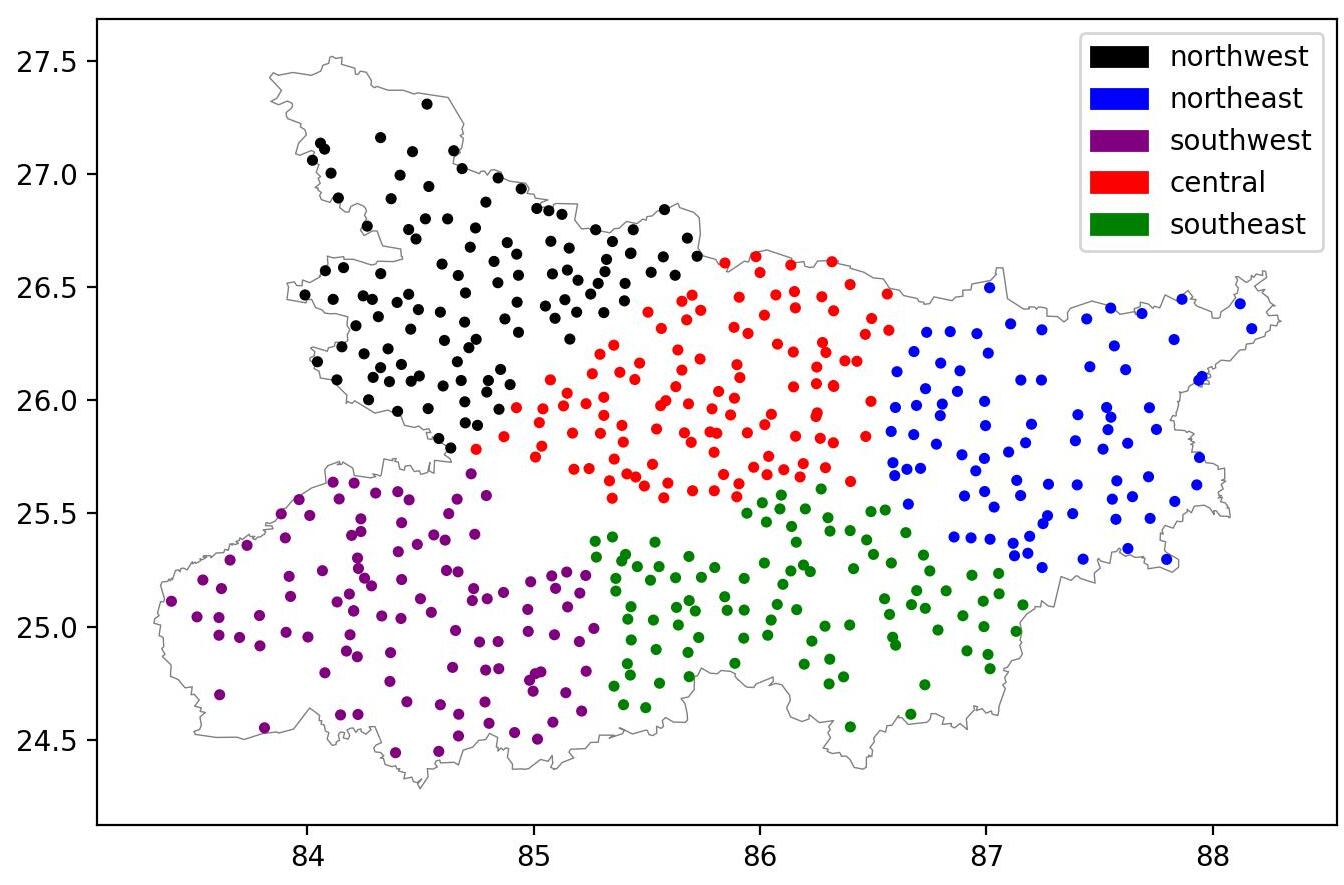}
    \caption{Distribution of AQM stations across the entire state of Bihar. The stations have been clustered region-wise.}
    \label{fig:2}
    \vspace{-5mm}
\end{figure}

\begin{figure*}[!htbp]
    \centering
    \begin{subfigure}{0.24\textwidth}
        \centering
        \includegraphics[width=\textwidth]{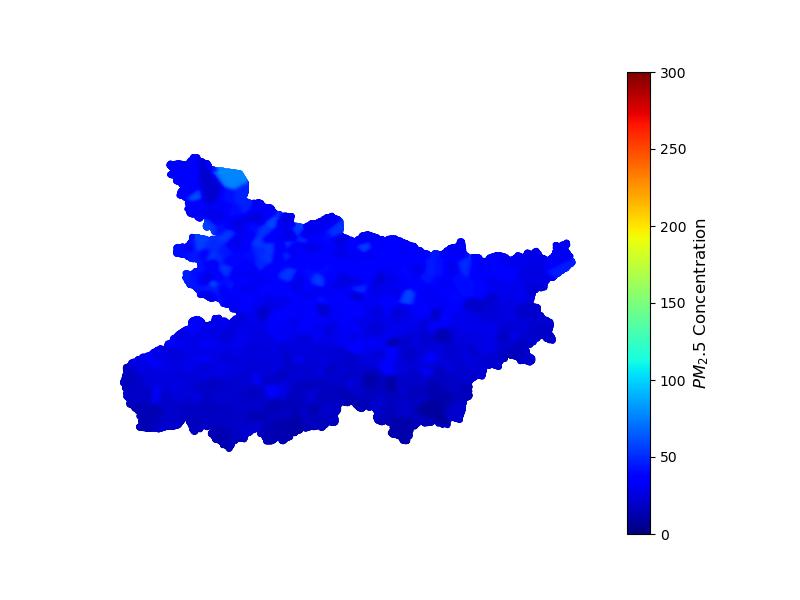}
        \label{fig:3a}
    \end{subfigure}
    \begin{subfigure}{0.24\textwidth}
        \centering
        \includegraphics[width=\textwidth]{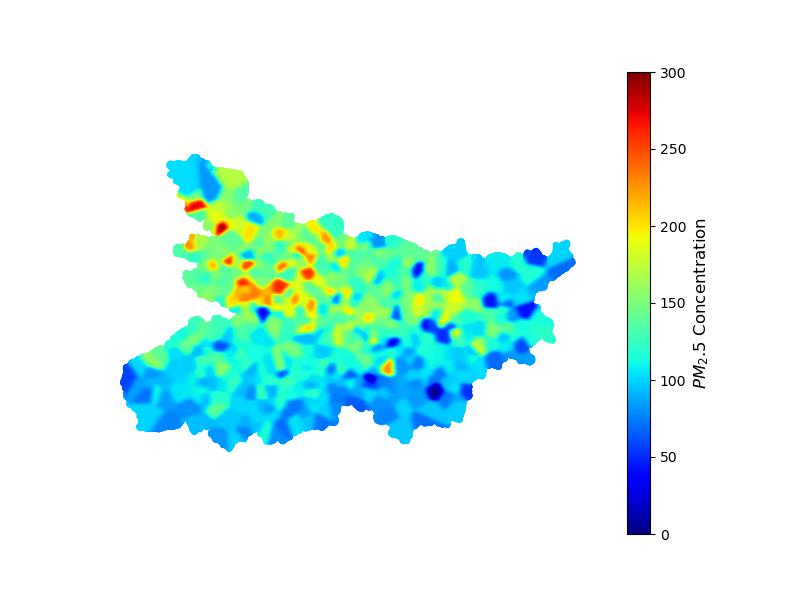}
        \label{fig:3b}
    \end{subfigure}
    \begin{subfigure}{0.24\textwidth}
        \centering
        \includegraphics[width=\textwidth]{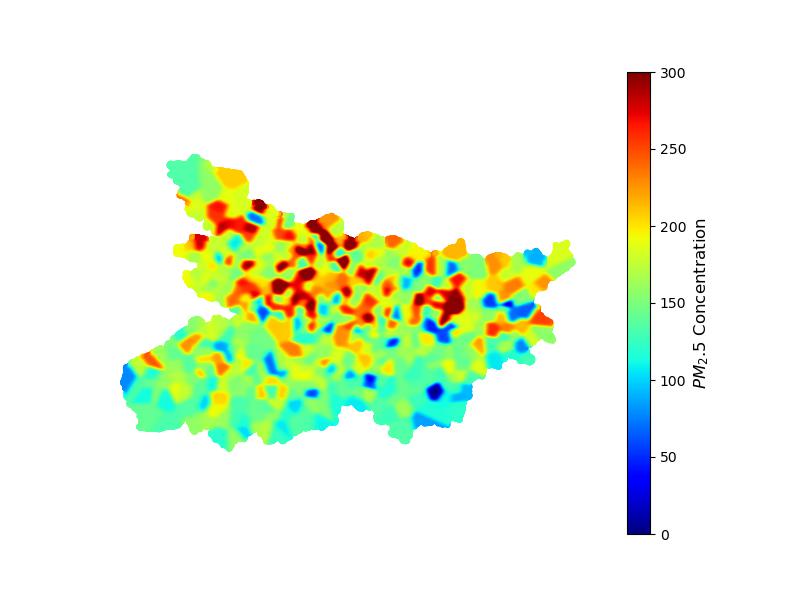}
        \label{fig:3c}
    \end{subfigure}
    \begin{subfigure}{0.24\textwidth}
        \centering
        \includegraphics[width=\textwidth]{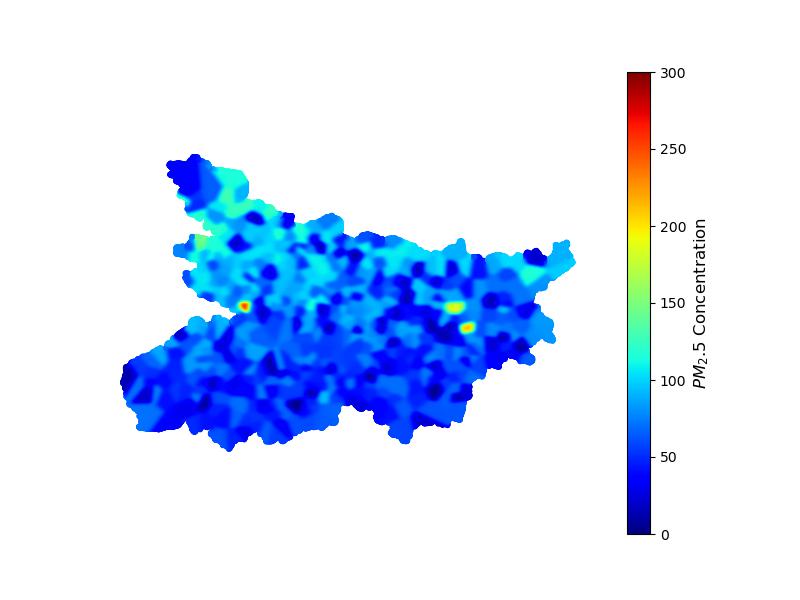}
        \label{fig:3d}
    \end{subfigure}
    \vspace{-2em}
    \caption{Spatial Distribution of mean PM$_{2.5}$ concentration for Bihar across seasons. From Left to Right: JJAS, ON, DJF, MA}
    \label{fig:3}
\end{figure*}

As stated in Section \ref{sec:method-dataset}, we collected a comprehensive dataset spanning an entire year (from 2023-05-01 to 2024-04-30) from 511 AQM stations (nodes), which practically covers most of the region of Bihar. To gain a better understanding of our model's performance, we have subdivided the AQM stations into different regions, as depicted in Figure \ref{fig:2}. This allows us to assess its effectiveness not only on a larger scale but also within specific regions. The AQM stations provide meteorological updates on an hourly basis. To investigate model capabilities, we temporally split the data into training, validation, and test sets, as mentioned in Table \ref{tab:exp}, since the PM$_{2.5}$ concentrations exhibit strong seasonality. Figure \ref{fig:3} illustrates the spatial distribution of mean PM$_{2.5}$ concentrations across the entire state of Bihar during four different seasons, namely: June-July-August-September (JJAS), October-November (ON), December-January-February (DJF), and March-April (MA). Specifically, the ON and DJF seasons show clear spikes in the Northwest, Central, and Northeast regions. The main causes of this phenomenon include winter inversion, dust storms, agricultural fires, and other pertinent aspects \cite{article}. It is imperative to include some of these months in the training dataset, as the PM$_{2.5}$ distribution during this period significantly differs from that of other seasons.

In addition to the Bihar dataset, we leverage the dataset employed by Wang et al. \cite{Wang_2020}, to not only demonstrate the robustness of our model across diverse regions, but also to showcase our model's capabilities in capturing seasonal trends over an extended time-frame spanning multiple years. The dataset covers the most polluted regions in China, covering a network of 181 cities (AQM stations), spanning 4 years (from 2015-01-01 to 2018-12-31), wherein the meteorological updates are provided every 3 hours.

\begin{table}
  \caption{Experiment Details}
  \label{tab:exp}
  \begin{tabular}{ll}
    \toprule
    \textbf{Parameter} & \textbf{Value}\\
    \midrule
    Number of Stations (L) & $511^{*}$, $181^{\dagger}$\\
    Update (hours) & $1^{*}$, $3^{\dagger}$\\
    Distance Threshold (km) & $5^{*}$, $300^{\dagger}$\\
    Haze Threshold ($\mu g/m^{3}$) & $100^{*}$, $75^{\dagger}$\\
    History Length (H) & 24, 48 hours\\
    Forecast Length (F) & 12, 24 hours\\
    Training Set $^{*}$ & 01/05/2023 - 31/12/2023\\
    Validation Set $^{*}$ & 01/01/2024 - 29/02/2024\\
    Test Set $^{*}$ & 01/03/2024 - 30/04/2024\\
    Training Set $^{\dagger}$ & 01/01/2015 - 31/12/2016\\
    Validation Set $^{\dagger}$ & 01/01/2017 - 31/12/2017\\
    Test Set $^{\dagger}$ & 01/01/2018 - 31/12/2018\\
  \bottomrule
\multicolumn{2}{r}{$^{*}\,:$ Bihar; $\dagger\,$: China} \\
\end{tabular}
\end{table}

\subsection{Baselines}

We compare our proposed method AGNN\_GRU with a number of high-performing baselines. It is important to highlight that these baselines are not only specific instances of our more general architecture but also represent special cases of several state-of-the-art models for spatio-temporal forecasting, including those proposed in ~\cite{GAO2021101150, TENG2024143542}. Including these baselines in our experiments thus serves as an ablation study, enabling us to thoroughly assess the contributions of the individual components within our proposed model. In our comprehensive ablation study, we systematically altered or removed certain components from our proposed model, which resulted in the following methods, which we subsequently compare with as strong baselines

\begin{itemize}
\item \textbf{GRU} is a vanilla sequence-to-sequence architecture that focuses only on temporal dependencies. This variant lacks a graph neural network in its encoder architecture and sends the last state of the encoder to the first state of the decoder. The concept of attention is not applied in this context.

\item \textbf{GC\_GRU} is an enhanced sequence-to-sequence architecture that incorporates only a basic undirected graph with binary edge weights, determined by distance threshold. This variant utilizes GCNConv \cite{kipf2017semisupervisedclassificationgraphconvolutional} as the message-passing layer.\\

$
    w_{ij} = \begin{cases}
        1 & d_{ij} \leq \text{distance threshold}\\
        0 & \text{otherwise}
    \end{cases}\\
$

\item \textbf{WGC\_GRU} is a more sophisticated sequence-to-sequence architecture featuring undirected weighted graph construction. The edge-weights are computed as follows:\\

$
    w_{ij} = \begin{cases}
        \frac{d_{min}}{d_{ij}} & d_{ij} \leq \text{distance threshold}\\
        0 & \text{otherwise}
    \end{cases}\\
$

This variant utilizes GraphConv \cite{morris2021weisfeilerlemanneuralhigherorder} as the message-passing layer.

\item \textbf{GNN\_GRU} is a sequence-to-sequence architecture utilizing the graph construction outlined in Section \ref{sec:method-architecture} but without implementing the attention mechanism.
\end{itemize}

As stated earlier, the method proposed in~\cite{Wang_2020} also consider modeling spatial-diffusion; however, they assume that the node attribute for future time-steps are available from another forecasting method. Therefore, their method is not directly comparable with ours because of the inherent differences in the problem setting, and is thus omitted from comparison.

We also do not include other classical methods such as ARIMA and others as they have been shown in prior works to be outperformed by the above baselines.

\begin{table*}[!htbp]
  \small
  \begin{tabular}{llllllll}
    \toprule
    \textbf{Model} & \textbf{Loss} & \textbf{RMSE} & \textbf{MAE} & \textbf{Spearman R} & \textbf{CSI (\%)} & \textbf{POD (\%)} & \textbf{FAR (\%)}\\
    \midrule
    AGNN\_GRU & \textbf{0.22 $\pm$ 0.01} & \textbf{34.04 $\pm$ 1.30} & \textbf{29.77 $\pm$ 1.08} & \textbf{0.77 $\pm$ 0.01} & \textbf{48.27 $\pm$ 0.62} & 76.81 $\pm$ 2.50 & 43.42 $\pm$ 1.92\\
    GNN\_GRU & 0.24 $\pm$ 0.02 & 35.22 $\pm$ 1.93 & 30.50 $\pm$ 1.94 & 0.73 $\pm$ 0.01 & 47.63 $\pm$ 0.56 & 72.56 $\pm$ 3.42 & 41.73 $\pm$ 2.67\\
    WGC\_GRU & 0.24 $\pm$ 0.02 & 35.37 $\pm$ 1.88 & 30.66 $\pm$ 1.75 & 0.74 $\pm$ 0.01 & 48.10 $\pm$ 0.61 & 75.02 $\pm$ 3.94 & 42.50 $\pm$ 3.13\\
    GC\_GRU & 0.23 $\pm$ 0.01 & 34.42 $\pm$ 1.07 & 29.83 $\pm$ 1.34 & 0.72 $\pm$ 0.02 & 47.37 $\pm$ 1.44 & 69.60 $\pm$ 3.99 & \textbf{40.16 $\pm$ 1.32}\\
    GRU & 0.28 $\pm$ 0.01 & 38.58 $\pm$ 0.90 & 33.61 $\pm$ 0.88 & 0.73 $\pm$ 0.01 & 46.94 $\pm$ 0.44 & \textbf{77.54 $\pm$ 1.52} & 45.64 $\pm$ 1.16\\
  \bottomrule
\end{tabular}
  \small
  \begin{tabular}{llllllll}
    \toprule
    \textbf{Model} & \textbf{Loss} & \textbf{RMSE} & \textbf{MAE} & \textbf{Spearman R} & \textbf{CSI (\%)} & \textbf{POD (\%)} & \textbf{FAR (\%)}\\
    \midrule
    AGNN\_GRU & 0.29 $\pm$ 0.01 & 42.56 $\pm$ 1.22 & 37.38 $\pm$ 1.23 & \textbf{0.73 $\pm$ 0.01} & \textbf{43.42 $\pm$ 0.58} & \textbf{83.85 $\pm$ 1.54} & 52.60 $\pm$ 1.15\\
    GNN\_GRU & 0.33 $\pm$ 0.02 & 45.79 $\pm$ 1.84 & 39.56 $\pm$ 1.64 & 0.72 $\pm$ 0.01 & 42.71 $\pm$ 0.99 & 83.79 $\pm$ 1.38 & 53.42 $\pm$ 1.55\\
    WGC\_GRU & 0.30 $\pm$ 0.01 & 43.37 $\pm$ 0.98 & 37.00 $\pm$ 0.95 & 0.70 $\pm$ 0.01 & 43.16 $\pm$ 0.32 & 78.19 $\pm$ 1.76 & 50.90 $\pm$ 1.07\\
    GC\_GRU & \textbf{0.27 $\pm$ 0.02} & \textbf{40.71 $\pm$ 2.07} & \textbf{34.32 $\pm$ 2.10} & 0.68 $\pm$ 0.01 & 43.14 $\pm$ 0.58 & 73.29 $\pm$ 3.22 & \textbf{48.26 $\pm$ 2.27}\\
    GRU & 0.29 $\pm$ 0.02 & 41.44 $\pm$ 1.81 & 35.90 $\pm$ 1.71 & 0.67 $\pm$ 0.02 & 42.21 $\pm$ 0.58 & 76.88 $\pm$ 4.69 & 51.47 $\pm$ 2.51\\
  \bottomrule
  \vspace{2mm}
\end{tabular}
\vspace{-1em}
\caption{\parbox[t]{0.75\linewidth}{Overall model performance on the Bihar dataset for 12-hour (top row) and 24-hour (bottom row) forecast settings. Best scores are in bold.}}
\vspace{-4mm}
\label{tab:bihar_stats}
\end{table*}

\begin{table*}[!htbp]
{\small
  \begin{tabular}{llllllll}
    \toprule
    \textbf{Model} & \textbf{Loss} & \textbf{RMSE} & \textbf{MAE} & \textbf{Spearman R} & \textbf{CSI (\%)} & \textbf{POD (\%)} & \textbf{FAR (\%)}\\
    \midrule
    AGNN\_GRU & \textbf{0.22 $\pm$ 0.01} & \textbf{15.10 $\pm$ 0.12} & \textbf{13.31 $\pm$ 0.11} & \textbf{0.78 $\pm$ 0.02} & 50.74 $\pm$ 0.16 & \textbf{63.93 $\pm$ 0.30} & 28.38 $\pm$ 0.90\\
    GNN\_GRU & 0.23 $\pm$ 0.01 & 15.30 $\pm$ 0.28 & 13.55 $\pm$ 0.29 & 0.78 $\pm$ 0.01 & \textbf{50.88 $\pm$ 0.15} & 63.72 $\pm$ 0.29 & \textbf{27.44 $\pm$ 0.77}\\
    WGC\_GRU & 0.23 $\pm$ 0.01 & 15.26 $\pm$ 0.09 & 13.47 $\pm$ 0.09 & 0.77 $\pm$ 0.01 & 50.19 $\pm$ 0.16 & 63.26 $\pm$ 0.32 & 28.05 $\pm$ 0.24\\
    GC\_GRU & 0.23 $\pm$ 0.01 & 15.20 $\pm$ 0.03 & 13.42 $\pm$ 0.03 & 0.77 $\pm$ 0.01 & 50.73 $\pm$ 0.38 & 62.79 $\pm$ 1.15 & 28.37 $\pm$ 0.20\\
    GRU & 0.24 $\pm$ 0.01 & 15.77 $\pm$ 0.07 & 14.01 $\pm$ 0.06 & 0.76 $\pm$ 0.01 & 50.12 $\pm$ 0.35 & 62.55 $\pm$ 0.55 & 29.98 $\pm$ 0.07\\
    \bottomrule
    \end{tabular}
}
{\small
  \begin{tabular}{llllllll}
    \toprule
    \textbf{Model} & \textbf{Loss} & \textbf{RMSE} & \textbf{MAE} & \textbf{Spearman R} & \textbf{CSI (\%)} & \textbf{POD (\%)} & \textbf{FAR (\%)}\\
    \midrule
    AGNN\_GRU & \textbf{0.30 $\pm$ 0.01} & \textbf{18.38 $\pm$ 0.09} & \textbf{15.70 $\pm$ 0.09} & \textbf{0.74 $\pm$ 0.01} & \textbf{48.16 $\pm$ 0.02} & \textbf{63.26 $\pm$ 0.23} & \textbf{28.92 $\pm$ 0.20}\\
    GNN\_GRU & 0.30 $\pm$ 0.01 & 18.52 $\pm$ 0.06 & 15.78 $\pm$ 0.06 & 0.73 $\pm$ 0.01 & 47.53 $\pm$ 0.77 & 59.43 $\pm$ 2.09 & 29.41 $\pm$ 0.42\\
    WGC\_GRU & 0.30 $\pm$ 0.01 & 18.57 $\pm$ 0.13 & 15.84 $\pm$ 0.14 & 0.73 $\pm$ 0.01 & 47.06 $\pm$ 0.23 & 58.72 $\pm$ 0.85 & 29.52 $\pm$ 0.56\\
    GC\_GRU & 0.30 $\pm$ 0.01 & 18.53 $\pm$ 0.15 & 15.79 $\pm$ 0.16 & 0.73 $\pm$ 0.01 & 47.00 $\pm$ 0.34 & 58.12 $\pm$ 0.58 & 29.58 $\pm$ 1.18\\
    GRU & 0.31 $\pm$ 0.01 & 18.98 $\pm$ 0.09 & 16.27 $\pm$ 0.09 & 0.73 $\pm$ 0.01 & 47.12 $\pm$ 0.38 & 58.54 $\pm$ 0.62 & 33.13 $\pm$ 0.26\\
    \bottomrule
    \vspace{2mm}
    \end{tabular}
}
\vspace{-1em}
\caption{\parbox[t]{0.75\linewidth}{Overall model performance on the China dataset for 12-hour (top row) and 24-hour (bottom row) forecast settings. Best scores are in bold.}}
\vspace{-3mm}
\label{tab:china_stats}
\end{table*}

\begin{figure*}[!htbp]
    \centering
    \begin{subfigure}{\textwidth}
        \centering
        \includegraphics[scale=0.35]{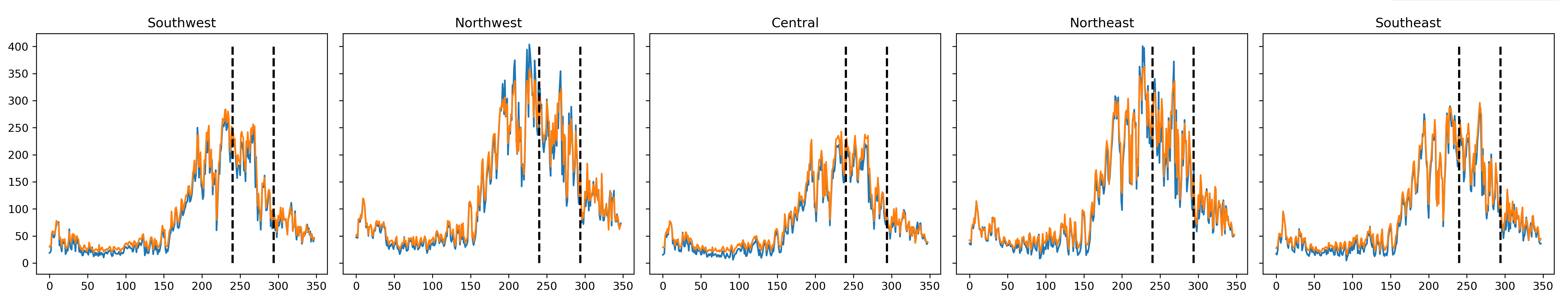}
        \label{fig:4a}
    \end{subfigure}
    \vspace{-2em}
    \begin{subfigure}{\textwidth}
        \centering
        \includegraphics[scale=0.35]{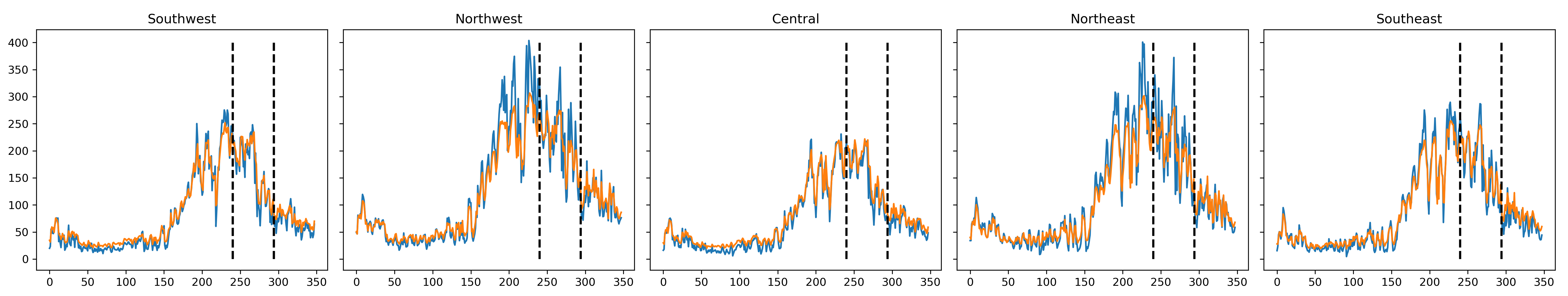}
        \label{fig:4b}
    \end{subfigure}
    \caption{Comparison of true (blue color) and predicted values (orange color) across different locations for Bihar dataset for 12-hour (top row) and 24-hour (bottom row) forecasting settings. Vertical dotted lines define the dataset partitions: training (left), validation (center), and test (right) sets.}
    \label{fig:4}
    \vspace{-1em}
\end{figure*}

\subsection{Experimental Settings}

We run experiments on a single NVIDIA A30 with 24G memory. As a pre-processing step, all the meteorological attributes, including the ground truth PM$_{2.5}$ concentrations have been normalized for better convergence. All the remaining details regarding experimental settings can be found in Table 4. After the complete training, we employ inverse standardization to get PM$_{2.5}$ concentrations back within the required range.
\subsubsection{Evaluation Metrics}
To evaluate the model performance, we consider test loss, Root Mean Squared Error (RMSE), and Mean Absolute Error (MAE) for evaluation where a smaller metric corresponds to better performance. Furthermore, we also report Spearman's rank correlation values, to check the correlation between our predictions and the ground truth. The model becomes more adept at analyzing spatio-temporal trends and adapting to changes as the value approaches 1, resulting in improved model performance. The RMSE is defined as $\text{RMSE} = \sqrt{\frac{1}{N} \sum_{i=1}^{N} (\, y_{i} - \hat{y}_{i})^{2}}$, the MAE is defined as $\text{MAE} = \frac{1}{N} \sum_{i=1}^{N} \big\lvert\, y_{i} - \hat{y}_{i} \, \big\rvert$, and  

\begin{align}
    \text{Spearman R} \, (\rho) &= \frac{\sum_{i=1}^{N} (y_{i} - \overline{y}_{i})(\hat{y}_{i} - \overline{\hat{y}}_{i})}{\sqrt{\sum_{i=1}^{N} (y_{i} - \overline{y}_{i})^{2}(\hat{y}_{i} - \overline{\hat{y}}_{i})^{2}}}
\end{align}

To check the model performance near a critical threshold (haze threshold), we also evaluate the Critical Success Index (CSI), Probability of Detection (POD), and False Alarm Rate (FAR), to account for abrupt changes in PM$_{2.5}$ concentration. To calculate CSI, POD, and FAR, we convert the real-valued predictions to binary labels, using haze threshold. The threshold is chosen as $100\,\mu g/m^{3}$ $(75\,\mu g/m^{3}$ for China), which is the average PM$_{2.5}$ concentration across all locations for the entire year. Following this threshold, we calculate hits (truth=1, prediction=1), misses (truth=1, prediction=0), and false alarms (truth=0, prediction=1). Note that the higher the CSI and POD values, the better the model performance. A smaller metric for FAR corresponds to a better performance. These metrics are defined as follows

\begin{align}
    \text{CSI} &= \frac{\text{hits}}{\text{hits} + \text{misses} + \text{false\_alarms}}\\
    \text{POD} &= \frac{\text{hits}}{\text{hits} + \text{misses}}\\
    \text{FAR} &= \frac{\text{false\_alarms}}{\text{hits} + \text{false\_alarms}}
\end{align}

The performance of each model is determined by averaging these metrics across all the locations. To get a fair estimate, we take five different initializations for each model, train on MSELoss, and subsequently report the mean and standard deviation of all the metrics.

\subsubsection{Hyperparameter Tuning}
In order to optimise the model's performance, we implement Bayesian search from wandb \cite{wandb} for 17 iterations for hyperparameter optimization. The hyperparameters that were optimised include learning rate [1e-5, 1e-2], regularisation parameter ($\lambda$) [5e-2, 5e-4], and embedding dimension (for the embeddings of the latitude-longitude based location features and timestamp features) $\{$8, 16$\}$. The model's performance on the validation set (MSE loss) was employed as the primary metric to determine the optimal configuration. After extensive cross-validation, the best-performing model used a learning rate of 1e-3, a lambda of 3e-3, and an embedding dimension of 8.

\vspace{-3mm}
\subsection{Results and Discussion}

Initially, we present the results from metrics mentioned in Section 4.3, and then demonstrate the characteristics of the proposed model (AGNN\_GRU), through careful analysis. To verify the performance of our proposed model, we present results in two experimental settings: (i) History Length of 24-hours with Forecast Length of 12-hours, and (ii) History Length of 48-hours with Forecast Length of 24-hours.

It is important to note that the dataset characteristics significantly influence the model's performance and the observed patterns. The dataset for Bihar spans only a single year, which limits the model's ability to capture long-term seasonality patterns. Consequently, we observe slightly more irregular patterns for this region. On the contrary, the dataset for China covers a span of 4 years, allowing the model to capture strong seasonality patterns. As a result, the results for China dataset are more coherent with our expectations, reflecting the impact of multi-year seasonal variations on the model's predictive capabilities.

\vspace{-2mm}
\subsubsection{12h Forecast Analysis}

As shown in Tables [\ref{tab:bihar_stats}, \ref{tab:china_stats}], our proposed model outperforms the baselines in almost all metrics. Specifically, our model obtains the lowest Normalized Test Loss, RMSE, and MAE, while simultaneously achieving the highest values for Spearman Rank Relation Coefficient and CSI (Bihar), and POD (China). The key metric to observe here is the Spearman Rank Relation Coefficient, as it shows that our model is the best at understanding the spatio-temporal trends since it has a value closest to 1. Also, CSI's (and POD's) high value indicates that our model captures abrupt trends better than any other model.
\vspace{-1em}
\subsubsection{24h Forecast Analysis}

Tables [\ref{tab:bihar_stats}, \ref{tab:china_stats}] summarizes the results in the 24-hour forecast setting. Our method gets the best results for the Spearman Rank Relation Coefficient, CSI, and POD. This means that our model is the best at detecting seasonal patterns as well as sudden changes in PM$_{2.5}$ concentrations near the haze threshold. However, our model did not achieve the lowest normalized test loss, which directly correlates with RMSE and MAE values, for the Bihar dataset. This might be attributed to the fact that our model is a lot more complex in terms of architecture than all the other models. Edge attributes contribute a lot to the increased number of parameters and, hence, the model complexity. Consequently, even though our model was able to capture the spatio-temporal trends the best, we believe that, given a dataset for a longer period, our model would converge also to a lower test loss and achieve improved values for all the statistics, as is evident from the China dataset, where our model performs the best in all the statistics.

Figure \ref{fig:4} illustrates our model's robust performance, not only at a macro level but also across diverse regional contexts. The ground truth values (and predictions) at a particular timestamp represent the mean of PM$_{2.5}$ concentrations (and predictions) of all the locations present within a particular localized region (northeast, northwest, central, southeast, southwest). We show the forecast values for the entire dataset period to demonstrate the model's effective performance. It can be clearly seen that not only is the model good at capturing trends in the training set but also in the validation set (DJF season), where there is a clear spike in PM$_{2.5}$ concentrations, and also in the test set. This demonstrates the model's capability to adjust to varying geographical and environmental conditions, suggesting its wide-scale applicability in forecasting.

Note that our model's performance, while robust, is constrained by the practical limitations of graph construction. The low distance threshold (5 km for Bihar) for edge creation, is necessitated by computational constraints since the graph edges scale quadratically as the number of nodes increases. This trade-off between model complexity and feasibility suggests the need to use more computing resources to model these spatial correlations even better.

\vspace{-2.5em}
\section{Conclusion}

In this paper, we study a significant problem of real-world interest: how to precisely forecast PM$_{2.5}$ concentrations across different locations. Due to the spatio-temporal nature of the problem, we designed AGNN\_GRU, a sequence-to-sequence architecture, coupled with a graph neural network, to capture spatial diffusion and factor in long-term dependencies. The experimental results illustrate the prospective efficacy of the proposed AGNN\_GRU technique in both short-term and long-term prediction tasks, as well as the model's generalizability across various geographies. The geography of China substantially differs with that of Bihar, as altitude above sea level is an important consideration in China. Notably, despite not relying on any pre-training, the model due to its relatively simpler architecture as compared to some of the other recent works on air-quality prediction, such as those based on transformers~\cite{liang2023airformer}, is still able to achieve impressive performance.

Future work could include some key areas to enhance the model's performance and applicability, as well as deploying these lightweight models on IoT devices for real-time forecasts. PM$_{2.5}$ concentrations exhibit significant seasonal variations, as depicted in Figure \ref{fig:3}, due to which, it becomes imperative to collect datasets, especially for longer periods rather than from more stations, is critical to understanding the recurring seasonal patterns in PM$_{2.5}$ concentrations. Such a dataset would likely improve both the quantitative and qualitative aspects of our model.\\


\bibliographystyle{ACM-Reference-Format}
\bibliography{sample-base}

\end{document}



\renewcommand{\shortauthors}{}

\begin{center}
\Huge{\textbf{Response to Reviewers' Comments}}
\end{center}

We thank the reviewers for their careful reading and helpful feedback. R\#3 has appreciated that the method does not require knowing the attributes at each location and also appreciated the use of the attention mechanism.  R\#4 has appreciated the novelty of capturing the relationship between time-series and exogenous variables through the combination of GNN and GRU, and the method’s applicability to real-world applications. R\#5 appreciated the work for designing and evaluating a model using a large-scale dataset that was collected in India.

There were some concerns and some suggestions from the reviewers.To address these, in the revised manuscript, we have incorporated the various suggestions from the reviewers (e.g., experiments on an additional large-scale PM2.5 dataset from China, and link to our codebase). In addition, in our response below, we discuss the concerns/questions from the reviewers and sincerely hope that our response (and the changes in the revised manuscript) addresses their concerns.